\documentclass[journal]{IEEEtran}

\usepackage{graphicx}
\usepackage{subfigure}
\usepackage{amsmath}
\usepackage{amssymb}
\usepackage{textcomp}
\usepackage{cite}
\usepackage{url}
\usepackage{color}
\usepackage{bbm}
\usepackage{mathtools}

\DeclareMathOperator*{\argmin}{arg\,min}  

\newcommand\Tstrut{\rule{0pt}{2.2ex}}
\newcommand\TstrutL{\rule{0pt}{4.4ex}}
\newcommand\Bstrut{\rule[-1.4ex]{0pt}{0pt}}   
\newcommand\BstrutL{\rule[-2.8ex]{0pt}{0pt}}   

\begin{document}
%
\title{Large-scale spatiotemporal photonic reservoir computer for image classification}
%
%
%

\author{Piotr~Antonik,
        Nicolas~Marsal,
        and~Damien~Rontani
        \thanks{P. Antonik, N. Marsal, and D. Rontani are with LMOPS EA 4423 Laboratory, CentraleSup\'elec, Universit\'e Paris-Saclay, F-57070 Metz, France, and Universit\'e de Lorraine, F-57000 Metz, France. emails : piotr.antonik@centralesupelec.fr; damien.rontani@centralesupelec.fr}%
\thanks{Manuscript received April 15, 2019; revised XXX xx, 2019.}}

%
%

\markboth{Journal of Special Topics in Quantum Electronics,~Vol.~X, No.~X, April~2019}%
{Shell \MakeLowercase{\textit{et al.}}: Bare Demo of IEEEtran.cls for IEEE Journals}
%

\IEEEspecialpapernotice{(Invited Paper)}

\maketitle

\begin{abstract}
  We propose a scalable photonic architecture for implementation of feedforward and recurrent neural networks to perform the classification of handwritten digits from the MNIST database. Our experiment exploits off-the-shelf optical and electronic components to currently achieve a network size of 16,384 nodes. Both network types are designed within the the reservoir computing paradigm with randomly weighted input and hidden layers. 
  Using various feature extraction techniques (\textit{e.g.} histograms of oriented gradients, zoning, Gabor filters) and a simple training procedure consisting of linear regression and winner-takes-all decision strategy, we demonstrate numerically and experimentally that a feedforward network allows for classification error rate of $\mathbf{1\%}$, which is at the state-of-the-art for experimental implementations and remains competitive with more advanced algorithmic approaches. 
  We also investigate recurrent networks in numerical simulations by explicitly activating the temporal dynamics, and predict a performance improvement over the feedforward configuration.
\end{abstract}


%
\IEEEpeerreviewmaketitle

\section{Introduction}
\IEEEPARstart{T}{he}  classification of static images is one of the central problem in Computer Vision in Machine Learning \cite{Duda2000,Cipolla2013} and has a broad range of applications such as in security and authentication with automatic facial recognition \cite{Liu2002,Ahonen2006} or object detection \cite{Ramanan2010}, in health sciences with the automatic analysis or segmentation of medical images \cite{Hames1992}, or intelligent character recognition for handwritten interpretation, such as automatic zip-code identification \cite{Lecun1990}.

Over the last decade, the level of accuracy of automatic image classification techniques has undergone a dramatic improvement coincidentally with to the breakthrough of deep-learning  and a particular class of neuro-inspired algorithms, known as convolutional neural networks (CNNs) \cite{lecun2015deep}. CNNs are multi-hidden-layered feed-forward neural network trained to extract hierarchical features from images and improve the recognition procedure thanks to their robustness to variability in viewpoint, scale, illumination conditions, and intra-class variations intrinsically present in images. 

Deep learning approaches based on CNNs have outclassed simpler classification systems, such as support vector machines (SVM) \cite{cortes1995support}. As a result, in $2015$, a CNN comprised of over $150$ layers has surpassed for the first time the accuracy of humans in classifying images from the ImageNet dataset \cite{ILSVRC15}, which is made of approximately 14 millions images distributed over more than $21,000$ different categories : It reached a classification error rate of $3.57\%$ to be compared to $5.1\%$ for humans. On smaller image datasets, such as the MNIST database made of 70,000 images of handwritten digits distributed over 10 different classes, deep-learning approaches reaches a classification error as low as $0.21\%$ \cite{lecun2015deep}. However, achieving high accuracy requires the training of a large number of hidden layers (\textit{i.e.} the fine tuning of millions of parameters). This is only possible by leveraging (very) large pre-labeled databases and the high-speed, massively parallelizable, computing power from graphical processing units (GPU), thus making the training procedure computationally demanding, usually intractable on regular computers, and energy intensive.

 An alternative approach to deep-learning for image classification uses (i) simple feature extraction techniques in combination with (ii) a recurrent neural networks (RNN) and (iii) a simplified training procedure with a limited number of learnable parameters. Reservoir Computing (RC) is a Machine Learning paradigm, which includes echo-state networks (ESN) and liquid-state machines (LSM), to design and train artificial recurrent neural networks \cite{jaeger2004harnessing,maass2002real}. Reservoir Computing exploits the transient response of a randomly-, sparsely-interconnected RNN made of nonlinear dynamical discrete-time nodes, which are subjected to randomly-weighted input signals. The output layer is made of (possibly) several outputs that realize linear combinations of the nodes' transient states. The training procedure for the Reservoir Computer is limited to the optimization of the output layer weights, which is usually obtained by simple linear (or ridge) regression \cite{lukosevicius2009reservoir}. This leads to a significant alleviation in the computational complexity of the training, because only the final unidirectional connections are optimized by comparison to the entire RNN in typical training procedures for artificial neural networks. Additionally, the reduced number of trainable parameters makes the reservoir computer efficient for classification even on small databases. 
 
 Because of its structural simplicity, reservoir computers have motivated many hardware implementations ranging from electronics \cite{appeltant2011information,Haynes2014} and spintronics \cite{Grollier2017,Grollier2019}, to photonics \cite{larger2012photonic,brunner2013parallel} with the goal of reaching ultra-fast processing speed with an energy consumption at least two order of magnitudes below that of a software-based RC running on a computer. Unprecedented classification speed have been recently achieved on the spoken-digit recognition task from the TIMIT46 data base with real-time processing speed ranging from 300,000 to 1,000,000 words analyzed per second using laser with optical feedback \cite{brunner2013parallel} and optoelectronic oscillators \cite{larger2017high}, respectively. 
 
 The challenge in photonics-based implementations of reservoir computing stems from the experimental difficulty to couple optically a large number (typically greater than $100$) of photonic devices together. To overcome this, the time-delay reservoir computer was introduced first with electronics systems \cite{appeltant2011information} and then with various optical/optoelectronic feedback configurations \cite{brunner2013parallel,larger2012photonic,larger2017high,paquot2012optoelectronic,Uchida2018,duport2012all} with various applications notably in optical communications \cite{Argyris2018}. The principle relies on a single nonlinear dynamical node subjected to time-delay feedback, where virtual nodes are located. This approach allows theoretically to increase the size of the network linearly with the length of the feedback loop but at the expense of the processing speed decreasing because data cannot be fed to the systems faster than the time-delay to get all the virtual nodes subjected to the data \cite{Soriano2017}.
 
To overcome this inherent drawback of the time-delay approach, true spatiotemporal photonic reservoir computers have been proposed : (i) using photonic integrated circuits (PICs) but limited in scale to a few tens of nodes due to high losses \cite{vandoorne2014experimental,Katumba2018,katumba2019neuromorphic}, which is inadequate for tackling complex tasks such as image classification or (ii) using free-space optics with a number of nodes reaching up to 2,500 in the most recent experiments \cite{bueno2018reinforcement}, thus making possible for such architecture to solve more complex tasks. 

The paper is organized as follows. 
First, in Section \ref{sec:rc} we present the basics of reservoir computing, provide a description of our experimental architecture for a large-scale photonic reservoir computer implementation exceeding 10,000 neurons, and derive the corresponding physical model used in our numerical simulations. Then, in Section \ref{sec:mnist}, we introduce the particular task of handwritten digit recognition based on the MNIST database and how our reservoir computer is adapted to solve it. We also describe the various feature extraction strategies (\textit{i.e.} histograms of oriented gradients, zoning, and Gabor filters) to generate the most relevant information from the images prior to their injection in the photonic recurrent neural network. 
Finally, we devote the Section \ref{sec:res} to the discussion of our results, where we compare three conceptually different modes of operation for our reservoir computer to successfully classify handwritten digits : the feedforward mode of operation, the recurrent mode with echo, and the recurrent mode with data-splitting. We analyze and compare the performance of these different modes of the reservoir computer when paired with the feature extraction layer. Finally, Section \ref{sec:ccl} concludes the paper and summarizes our contributions.

\section{Spatiotemporal Photonic Reservoir Computer}
\label{sec:rc}

In this section, we describe the basics principles of reservoir computing and how we realize and model our experimental spatiotemporal photonic reservoir computer with up to 16,384 nodes.

\subsection{Principles of reservoir computing}
\label{subsec:rc}

A reservoir computer consists of a RNN with $n\in\mathbb{N}$ discrete-time nonlinear randomly interconnected dynamical systems, which is mathematically described by the following state equation 
\begin{equation}
  \mathbf{x}(k+1) = \mathbf{f}_{\text{NL}} \left( W_{res} \mathbf{x}(k) + W_{in} \mathbf{u}(k) \right),
  \label{eq:rcevo}
\end{equation}
where $\mathbf{x}(k)\in \mathbb{R}^n$ is the state vector of the RNN at discrete time $k\in\mathbb{N}$; $\mathbf{f}_{\text{NL}}$ is a nonlinear vector flow mapping the state space $\mathbb{R}^n$ to itself. $W_{res}\in\mathbb{R}^{n\times n}$ is the adjacency matrix of the RNN describing the topology of the weighted interconnections between various neurons. The spectral radius of the matrix $W_{res}$ is usually taken in the range $0.8-1.1$ to guarantee the overall stability of the RNN and the fading memory property, one of the necessary condition on the RNN to be used in the framework of reservoir computing \cite{jaeger2001echo,jaeger2004harnessing}. $W_{in}\in\mathbb{R}^{p\times n}$ is the input matrix representing the weighted interconnections between the input layer and the RNN, also known as the \textit{input mask}. Similarly to $W_{res}$, its  weights are randomly distributed with zero mean. Lastly, $\mathbf{u}(n)\in\mathbb{R}^{p}$ is the input vector. 

When subjected to time-varying input data, the state vector of the RNN undergoes complex transient dynamics before returning to a stable quiescent state. These transient dynamics are wire-tapped and weighted to form an output vector $\mathbf{y}(n)$ described by the following linear relation
\begin{equation}
  \mathbf{y}(k) = W_{out} \mathbf{x}(k),
  \label{eq:rcout}
\end{equation}
where $W_{out}\in{n\times m}$ is the output matrix  representing the unidirectional interconnections between the RNN and the ouput layer. The elements of the output matrix, also called \emph{readout weights}, are the only learnable parameters of our reservoir computer architecture. They are trained using a convex optimization problem known as ridge regression (or Tikhonov regularization) \cite{tikhonov1995numerical} that reads :
\begin{equation}
   W_{out,opt} = \argmin_{W_{out}\in\mathbb{R}^{m\times n}}\|\mathbf{Y}_t-W_{out}\mathbf{X}\|^2 + \lambda\|W_{out}\|^2,
\end{equation}
where $\mathbf{Y}_{t}\in \mathbb{R}^{m\times k_{tr}}$ is the target matrix containing the $k_{tr}$ targets values for the $m$ outputs of the reservoir computer and $\mathbf{X}\in \mathbb{R}^{n\times k_{tr}}$ is the reservoir states matrix containing the RNN's transient states $\mathbf{x}(k)$ sampled during the training, when inputs vectors are fed to the reservoir during $k_{tr}$ time steps. $\lambda$ is a positive factor called ridge parameter used to control the norm of the readout weights by penalization; it is usually determined by cross-validation, when $\lambda=0$ then the ridge regression corresponds to linear regression. 

The resolution of the optimization problem to obtain the optimized readout weights in $W_{out,opt}$ is performed either off-line or online \cite{antonik2017online}, \emph{i.e.} either after all the training input data are injected in the reservoir, or in the course of the injection.

\subsection{Experimental photonic reservoir computer}
\label{sec:RCexp}

\begin{figure}[t!]
  \includegraphics[width=8.5cm]{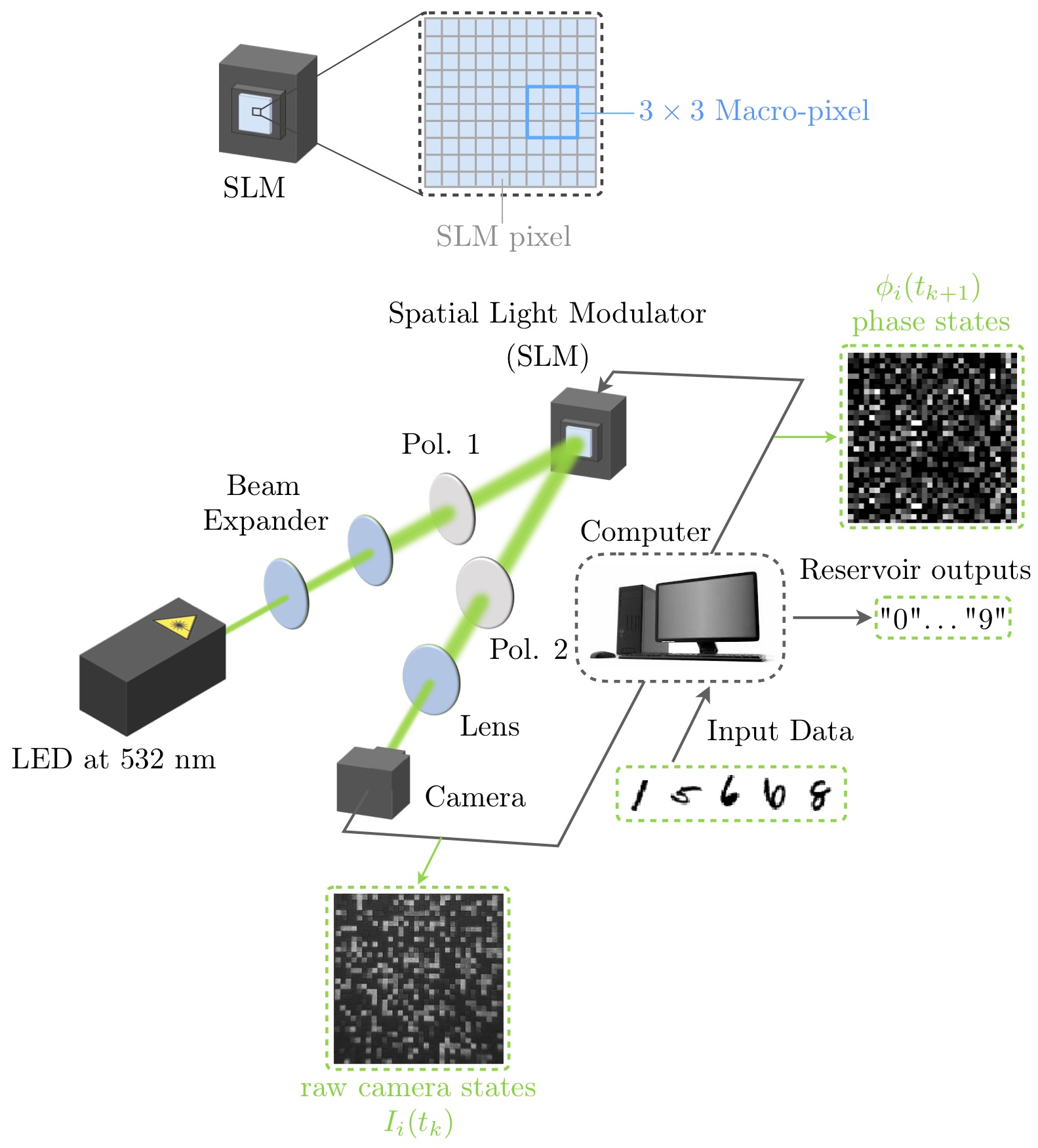}
  \caption{Scheme of the experimental setup composed of an optical arm linked to a computer. A collimated polarized green (532 nm) LED beam is reflected by the surface of the spatial light modulator (SLM). The SLM is imaged onto a camera through a polarizer (Pol{.} 2) and an imaging lens. Both the SLM and the camera are driven by a computer running a Matlab script. The latter generates the inputs from the MNIST database images, then computes the pixel values to be loaded on the SLM. Larger macro-pixels made of groups of individual SLM's pixels are used to train the reservoir in such a way to facilitate the separation on the raw camera images. The computer receives the data containing the reservoir states from the camera, computes the outputs and generates the output digits.}
  \label{fig:experiment}
\end{figure}

The proposed photonic network is depicted in Fig{.} \ref{fig:experiment} and its structure is inspired in parts from recent experiments in Refs{.} \cite{hagerstrom2012experimental,bueno2018reinforcement}. The setup consists of a photonic arm comprising a collimated incoherent monochromatic light source at $532$ nm (Thorlabs  M530L3), a set of two polarizers oriented at a $45^\circ$ angle with respect to the vertical axis, placed before and after a liquid-crystal on Silicon (LCoS), phase-only, spatial-light modulator (SLM) with a $512\times 512$ pixels with a $8$ bit resolution (Meadowlark P$512-0532$). This particular combination of polarizing optics and a SLM allows for the phase pattern imprinted in the transverse plan of the optical beam to be nonlinearly converted into an amplitude pattern and hence ensures an all-optical implementation of the nonlinear flow $\mathbf{f}_{NL}$ from Eq{.} (\ref{eq:rcevo}). Finally, the intensity pattern is imaged through a lens on a 1.3-Million-pixels high-speed camera with 10-bit resolution (Allied Vision Mako U130B).

The detected patterns are digitized and processed in the electronic arm of the setup. First, they are linearly combined on a computer (which could potentially be replaced by a DSP board), that effectively realizes the adjacency matrix of the network $W_{res}$ with full control over their choices. Then, input signals are added numerically after the masking operation with the $W_{in}$ matrix. Finally, these signals are fed back to the SLM controller, that changes the liquid-crystal orientation of each SLM's pixel at a given discrete time step and hence generates a new phase pattern based on the signals detected at the previous time step. This setup forms a dynamical photonic-based network, which we will be described mathematically as coupled maps in Sec. \ref{sec:RCmod}.

The size of the reservoir is essentially limited by the lowest resolution between that of the SLM and the camera. In our current setup, this means that, potentially, we could reach a network size of $512\times512 = 262,144$ nodes, where each pixel of the SLM, imaged with a $1:1$ ratio on the camera, could be considered as a physical node in our photonic network. However, we only use pixels located at the center of the SLM matrix due to constrains stemming from (i) inhomogeneous intensity distribution of the optical beam and (ii) small misalignment between the SLM and camera optical axes. To mitigate the effects of these experimental imperfections, we regroup neighbouring pixels together to form bigger square-shaped pixels (see Fig. 1(b)), latter referred to as \emph{macro-pixels} in this study, and becoming our physical nodes in the photonic network. The smallest macro-pixel effectively usable in our experience is $3\times 3$, which allows for $n=16,384$ macro-pixels available at the center of the SLM. This is currently our experimental upper-bound for the network size, but using a higher-resolution camera and SLM, and possibly smaller macro-pixel sizes with more stringent alignment conditions, our network could easily scale to hundreds thousands nodes.

The speed of the setup is defined by the time needed to compute the next SLM matrix from the raw camera image. This operation is deliberately performed in Matlab for greater flexibility of the experimental scheme. The system is capable of processing 2 images per second with a larger reservoir ($n=16,384$) and up to 7 frames per second with a small reservoir ($n=1,024$).
The system's speed limitation can be alleviated by replacing the computer with a dedicated digital signal processing (DSP) board, or a field-programmable gate array (FPGA) chip, capable of performing the matrix-products computations in real time (as in e.g. \cite{antonik2017online}).
Matrix multiplication can also be offloaded to fully parallel optics \cite{bueno2018reinforcement, psaltis1985optical}.
Furthermore, a dedicated computing unit (FPGA or DSP board) could also take care of the feature extraction stage, thus allowing real-time processing of the input images.

\subsection{Physical Modelling of the photonic reservoir computer}
\label{sec:RCmod}

In this section, we propose a phenomenological modelling of the dynamics of our photonic network.

We first model the optical arm and assume, for simplicity, that the optical field $\mathbf{E}_i(t_k)$ and phase pattern $\phi_i(t_k)$ are homogeneous and constant over the surface of the $i$-th macro-pixel during $\Delta t = t_{k+1}-t_{k}$, the sampling time of our electronic feedback loop. Furthermore, using only the macro-pixels close to the optical axis of the SLM, we also assume homogeneity of the optical field over the entire $n=16,384$ macro-pixels and hence $\mathbf{E}_i(t_k) = \mathbf{E}(t_k)$ and its constant value $\mathbf{E}(t_k)=\mathbf{E}_0$ during our experiment.

The incoherent, unpolarized light source is filtered by a linear polarizer oriented at $45^\circ$ with respect to the vertical axis, hence leading to the following expression $\mathbf{E}_{0} = E_{0}/\sqrt{2}[1,1]^T$. Each SLM's  macro-pixel is composed of liquid-crystal cells on top of a highly-reflective surface. The liquid crystal cells of the $i$-th macropixel can be considered as a birefringent material with its fast-optical axis along the vertical direction and with a programmable phase-shift $\phi_i(t_k)$. Hence, the Jones Matrix for the SLM's $i$-th macropixel is given by 
\begin{equation}
J_{SLM,i}(t_k) = \left(\begin{array}{cc}
     e^{i\phi_i(t_k)/2} & 0  \\
     0 & -e^{-i\phi_i(t_k)/2}
\end{array}\right).
\end{equation}
After reflexion on the SLM $i$-th macro-pixel and transmission through the second polarizer oriented at $45^\circ$ and described by the Jones matrix $J_{p,45}$, the output field is given by $\mathbf{E}_{out,i}(t_k) = J_{p,45}J_{SLM,i}(t_k)\mathbf{E}_{0}=\mathbbm{i}\frac{E_0}{\sqrt{2}}\sin\left(\frac{\phi_i(t_k)}{2}\right)\left[1,1\right]^T$.
The intensity detected by the camera and associated to the $i$-th macro-pixel hence reads
\begin{equation}
    I_i(t_k) = \|\mathbf{E}_{out,i}(t_k)\|^2 = I_0\sin^2\left(\frac{ \phi_i(t_k)}{2}\right),
\end{equation}
with $I_0 = \|\mathbf{E}_0\|^2$ the constant and uniform optical intensity emitted by the LED. 

The camera has a 10 bits resolution and detects a quantified version of the intensity of the $i$-th macro-pixel on the SLM that we will denote $\lfloor{I_i(t_n)}\rfloor_{10}$. We define the $n$-dimensional state vector of the photonic network from the macro-pixel intensities detected by the camera by:
\begin{equation}
\mathbf{x}(t_k) = \left[\lfloor {I_1(t_k)} \rfloor_{10},\dots,\lfloor {I_n(t_k)}\rfloor_{10}\right]^T.
\end{equation}

The 10-bits quantified intensities detected by the camera (corresponding to the SLM's macro-pixels) are linearly combined together and masked input data is added digitally. This operation on the states are then quantified over $8$ bits before being fed back to the SLM controller so that phase values of the SLM's macro-pixels $\phi_i(t_{k+1})$ for $i={1,\dots,n}$ are updated and so are the intensities $I_i(t_{k+1})$. The output of the reservoir are determined by combining linearly the intensities detected by the cameras. This leads to the following state and output equations :
\begin{eqnarray}
 \mathbf{x}(t_{k+1}) &=& \left\lfloor I_0 \sin^2\left(\left\lfloor W_{res}\mathbf{x}(t_k) + W_{in} \mathbf{u}(t_k) \right\rfloor_{8}\right)\right\rfloor_{10},\label{eq:evoPhotRC}\\
 \mathbf{y}(t_k)&=&W_{out}\mathbf{x}(t_k).\label{eq:outPhotRC}
\end{eqnarray}
Equation (\ref{eq:evoPhotRC}) should be understood as the nonlinear function $\lfloor I_0\sin^2\left(\lfloor \cdot \rfloor_{8}\right)\rfloor_{10}$ applied to each coordinate of the $n$-dimensional phase space. Equations (\ref{eq:evoPhotRC})-(\ref{eq:outPhotRC}) are the tailored version of Eqs. (\ref{eq:rcevo})-(\ref{eq:rcout}) specific to our photonic implementation. This model will be later used in our numerical simulation to assess the performance of the photonic reservoir computer.

\section{Reservoir Computing with Feature extraction for image classification}
\label{sec:mnist}

Feature extraction is a common approach in computer vision to enhance the classification system, here the photonic reservoir computer, by providing the most relevant information in images.  
In this section, we first introduce the approach adopted with reservoir computer to solve the MNIST handwritten digit recognition task and then present the five
feature extraction techniques 
that we tested in this study. 

\subsection{Resolution of the handwritten digit recognition task from the MNIST database with reservoir computing}

\begin{figure}[t]
  \centering
  \includegraphics[width=8.5cm]{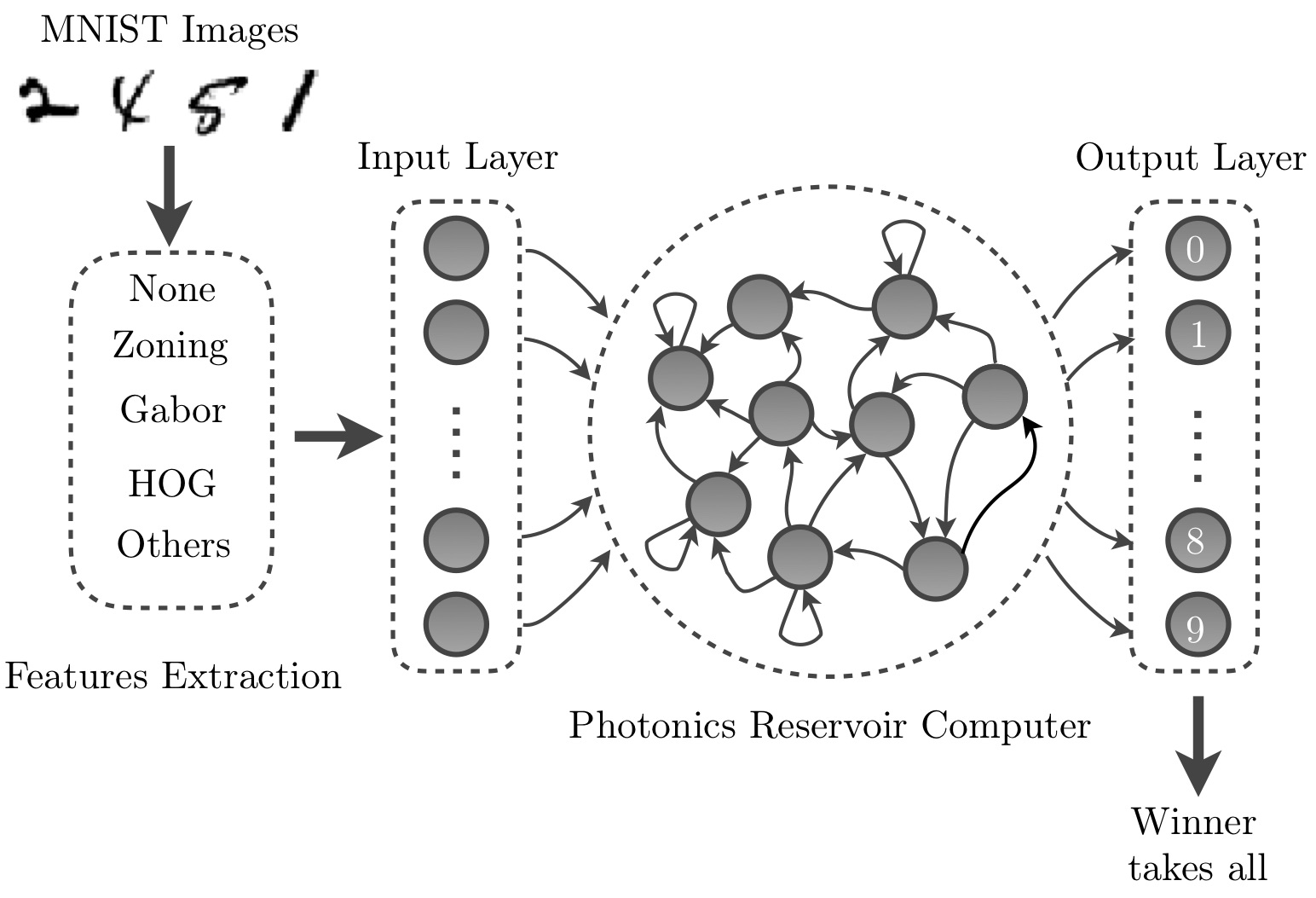}
  \caption{Handwritten digit recognition in the context of photonic reservoir computing with feature extraction. The images from the MNIST database undergo a feature extraction stage, where different algorithms under investigations are applied (or none, in the case of raw images). The resulting features are fed into the photonic reservoir computer with 10 binary output nodes, one for each digit. The nodes are trained to output 1 for the digit associated with the node and 0 for the other digits. The final classification is obtained by selecting the node with the maximum output, \textit{i.e.} the winner-takes-all decision strategy.}
  \label{fig:principle}
\end{figure}

The principle of the handwritten digit recognition in the context of reservoir computing with feature extraction is illustrated in Fig{.} \ref{fig:principle}.
In this work, we used the popular MNIST database\cite{lecun1998gradient}, publicly available online, which contains $70,000$ images of handwritten digits from 0 to 9. All images have been normalized to fit into a $28\times28$ pixels bounding box, anti-aliased, and converted into gray-scale levels.

The input images can be pre-processed in several ways to the input layer of the reservoir computer, depending on the type of feature extraction presented in Sub-section \ref{subsec:feat}, and it will be shown in Section~\ref{sec:res} that this choice impacts the classification error significantly. 

The output layer is made of $10$ binary outputs, introduced to recognize the 10 different class of digits. Hence, each binary output is trained to give a ``1'' for an image of the digit it is associated to and ``0'' for all other digits. The winner-takes-all decision strategy is used to classify each image based on the binary output with the highest value.

Furthermore, the photonic reservoir computer can have different modes of operation and processes  the masked input data differently depending on how the latter is fed into the reservoir. 
In this study, we consider three different modes of operation.

\emph{Full-image feedforward mode} -- In this mode, one full MNIST image is sent to the reservoir at each timestep $t_k$. To avoid cross-talk between consecutive images caused by the internal memory of the system, the reservoir should be exploited as memoryless map with $W_{res}=0_{n\times n}$. Therefore, the state equation of the reservoir reduces to:
\begin{equation}
    \mathbf{x}(t_{k+1})=\left\lfloor I_0 \sin^2\left(\left\lfloor W_{in} \mathbf{u}(t_k) \right\rfloor_{8}\right)\right\rfloor_{10}.
\end{equation}
The results obtained with this approach will be presented in Sec. \ref{subsec:res:ff}.

\emph{Full-image recurrent mode} -- A more advanced mode of operation is to present a full MNIST image at a given timestep $t_k$ and let the reservoir process the information during $k_e$ additional timesteps until the system relaxes into a quiescent state. Then, a new image can be sent at $t_{k+ke+1}$. In this mode of operation, $ W_{res}\ne 0_{n\times 0}$ and the state equation of the reservoir is given by Eq{.}~(\ref{eq:evoPhotRC}). We will discuss this approach further in Sec. \ref{subsec:res:rc}.

\emph{Column-wise recurrent mode} $-$ The third mode of operation for the reservoir is inspired by Ref{.}\cite{schaetti2017echo} and consists of dividing each MNIST image into $28$ columns. The image is then presented at a rate of one column per time step $k$, so that it takes 28 timesteps to feed a single image to the reservoir. In this mode of operation, the dynamics of the network is described by Eq{.}~(\ref{eq:evoPhotRC}) with $W_{res}\ne 0_{n\times 0}$. The underlying idea here is to transform the spatial correlations naturally present in images into temporal correlations in the dynamical states of the reservoir and thus exploit the internal memory of the system created by the recurrence of the network. The disadvantage, however, is a processing speed divided by $28$ compared to the first mode of operation (full-image feedforward mode). The results for this mode of operation will be presented in Section \ref{subsec:res:col}.

\subsection{Feature extraction approaches}
\label{subsec:feat}

We implemented and compared five feature extraction techniques, also considered in \cite{bahi2015robust}. 
The three approaches that gave the best results are presented in this section, and their results will be discussed in Sec. \ref{sec:res}. 
Table \ref{tab:feat} summarizes these techniques and their respective input dimensionalities (i.e. the number of features).
The two other approaches that did not yield acceptable classification errors are only briefly discussed in the end of this section.

\begin{table}[!htbp]
  \centering
  \begin{tabular}{lc}
     \hline
    Feature extraction method & Input dimensionality \Tstrut\Bstrut\\
    \hline
    Raw images & 784 \TstrutL\\
    Zoning 2 & 196 \\
    Zoning 4 & 49 \\
    Gabor filters & 40 -- 1152 \\
    HOG & 324 -- 1296 \BstrutL\\
        \hline
  \end{tabular}
  \vspace{0.25cm}
  \caption{Feature extraction techniques investigated in this work}
  \label{tab:feat}
\end{table}

\subsubsection{Raw images}
\label{subsec:feat:raw}

We used raw images -- \textit{i.e.} without any pre-processing -- as a benchmark for the feature extraction methods considered below. 
In this particular case, grey-scale values of the pixels are used as inputs to the reservoir computer, and the dimensionality of the input is $28 \times 28 = 784$. An example of raw image is shown in Fig{.}~ Fig. \ref{subfig:feat:raw}.

\subsubsection{Zoning}
\label{subsec:feat:zon}

Introduced in Ref{.}~\cite{hussain1972results}, it is the simplest feature extraction technique considered here. It is a statistical region-based approach, where the image is divided into smaller zones and pixel densities are computed in each one. In other terms, this method consists of a combination between a convolution of the image by a filter, followed by pooling a single value from each zone of the image.

Here, we consider two variants of this approach, designed specifically for the MNIST database: \emph{Zoning 2} with $2\times2$-pixel zones, and \emph{Zoning 4} with larger $4\times4$-pixel zones. 
The filters are $2\times2$ or $4\times4$ matrices filled with $1$s, thus giving a non-normalized average pixel values for each zone. This technique allows to reduce the input-dimensionality by ``dropping'' $3$ out of $4$, or $15$ out of $16$ pixels for Zoning 2 and Zoning 4, respectively. Consequently, the input dimensionality is $196$ for Zoning 2, and $49$ for Zoning 4. The resulting inputs to the reservoir computer are illustrated in Figs{.} \ref{subfig:feat:zon2} and \ref{subfig:feat:zon4}.

\subsubsection{Gabor filters}
\label{subsubsec:feat:gab}
They are linear filters mainly used for texture analysis \cite{jain1991unsupervised} and have been also used to model the receptive field profiles of simple cells in mammalian visual cortex  \cite{bovik1990multichannel,turner1986texture,perry1989segmentation,tan1990texture}. Therefore, pre-processing images with Gabor filters could be compared, to a certain extent, to perception in the human visual system.
The 2D-impulse response $h(x,y)$ of a Gabor filter is defined by the product of a Gaussian function with a sine/cosine  function. There are either complex-valued Gabor filters \cite{bovik1990multichannel} or pairs of Gabor filters with quadrature-phase relationship \cite{turner1986texture,perry1989segmentation,tan1990texture}. In this work, we use real-valued, even-symmetric Gabor filters\cite{jain1991unsupervised,malik1990preattentive}, which  impulse response is given by :
\begin{equation}
  h(x,y) = \exp \left( - \frac{x_\theta^2}{2\sigma^2_x} - \frac{y_\theta^2}{2\sigma_y^2} \right) \cos \left( \frac{2\pi}{\lambda} x_\theta \right),
  \label{eq:gabor}
\end{equation}
with
\begin{align}
    x_\theta & = x \cos (\theta) + y \sin (\theta), \\
    y_\theta & = y \cos(\theta) - x\sin(\theta),
\end{align}
where $\lambda$ is the wavelength of the cosine function with orientation $\theta$ with respect to the $x$-axis and $\sigma_{x,y}$ are the space constants of the Gaussian envelope along the $x,y$ axes, respectively. 
A Gabor filter seeks for a specific frequency content in a direction $\theta$ by yielding higher values in locations visually similar to the filter itself after convolution with the image. The dimensionality of the resulting features depends on (i) the number of filters, differentiated by their lengths and orientations, and (ii) the computation of the local energy, that will be discussed in Section~\ref{subsubsec:res:ff:gab}.

\subsubsection{Histograms of oriented gradients (HOG)}
\label{subsubsec:feat:hog}

This algorithm is widely used in computer vision for object detection and locati    on \cite{bahi2015robust} in static images. It was originally proposed in Ref{.}~ \cite{dalal2005histograms} and is based on scale-invariant features transform (SIFT) descriptors \cite{lowe2004distinctive}. The HOG descriptors are obtained by creating histograms on the magnitude of images gradient $m(i,j)=(D_x(i,j)^2+D_y(i,j)^2)^{1/2}$ depending on their orientation $\theta(i,j)=\arctan(D_y(i,j)/D_x(i,j))$ with $D_x(i,j)$ and $D_y(i,j)$ the horizontal and vertical approximate gradients evaluated at pixel $(i,j)$. These gradients are computed using the two spatial filters $G_x = \left[ -1, 0, 1 \right]$ \text{and} $G_y = \left[ -1, 0, 1 \right]^T$. 

\begin{figure}[!t]
  \centering
  \subfigure[]{\includegraphics[width=0.15\textwidth]{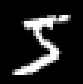}\label{subfig:feat:raw}}
  \subfigure[]{\includegraphics[width=0.15\textwidth]{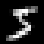}\label{subfig:feat:zon2}}
  \subfigure[]{\includegraphics[width=0.15\textwidth]{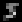}\label{subfig:feat:zon4}}
  \subfigure[]{\includegraphics[width=0.15\textwidth]{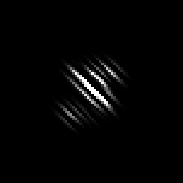}\label{subfig:feat:gabor}}
  \subfigure[]{\includegraphics[width=0.15\textwidth]{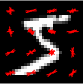}\label{subfig:feat:hog}}
  \caption{Illustration of \textbf{(a)} raw MNIST image, and the feature extraction methods, applied to it: \textbf{(b)} Zoning 2, \textbf{(c)} Zoning 4, \textbf{(d)} Gabor with $\lambda=3$ and $\theta=22.5^\circ$ orientation capturing the middle stroke of the digit, and \textbf{(e)} HOG features (red) computed with $7\times7$ cells and 9 orientation bins, superimposed on top of the source image.}
  \label{fig:feat:zon}
\end{figure}

A typical histogram consists of $9$ bins in the $[0^\circ,180^\circ]$ interval. It is calculated on $8\times 8$-pixel cells within the image. This ensures a dimensional reduction from $64$ gray-scale intensity values to $9$ cumulative frequencies values corresponding to the so-called HOG descriptors. To make them robust to lighting inhomogeneities, the histograms are block-normalized over larger regions of pixels within the image. 
Similarly to Zoning, the number of features depends on the size of the image and the cell, the number of bins, and the block-normalisation. In the case of the MNIST database, the total number typically ranges between 234 and 1296. This will be further discussed in Sec. \ref{subsubsec:res:ff:hog}. 
Figure \ref{subfig:feat:hog} illustrates the resulting gradients superimposed on top of an example image from the MNIST dataset. 


  
%
%

\subsection{Other methods}
\label{subsec:feat:misc}


The following two feature extraction methods were investigated in preliminary simulations, but were not included in the results (Sec. \ref{sec:res}) since they performed worse than the raw images. Moreover, both methods require the distinction between foreground and background pixels, and are thus more complex to implement on real-life images.

\subsubsection{Projection histograms}

Projection histograms (vertical and horizontal) were considered in \cite{bahi2015robust}. This approach consists in counting the number of white foreground pixels in each column or row, respectively. We used both histograms simultaneously -- that is, 56 features for each image -- and obtained classification errors of order of $12\%$.

\subsubsection{Distance profiles}

This descriptor calculates the pixel-distance between the border of the image and the first pixel of the foreground (i.e. the digit)\cite{siddharth2011handwritten}. Based on the four borders of the image, four profiles can be considered: left, right, top, and bottom. In this study, we investigated the combination of all four profiles, since it yields a descriptor with the most information on the image, encoded into 112 features. This approach works better than the projection histograms above, but we could not reduce the classification error below circa $10\%$.

\section{Results}
\label{sec:res}

The results of this study are divided into three sections, based on how the MNIST images are processed by the reservoir computer (see Sec. \ref{sec:mnist}).

\subsection{Full-image feedforward mode}
\label{subsec:res:ff}

Presenting one full MNIST image per time step $k$ is the simplest approach considered here. Besides the simplicity, it has the advantage of (1) being compatible with feature extraction techniques, presented in Sec. \ref{subsec:feat}, and (2) not increasing the number of processing time steps, since it is equal to the size of the MNIST database. The downside, however, is that it does not exploit the temporal dynamics of the reservoir computer. In fact, since the individual MNIST images are independent from each other, the inputs to the reservoir lack the temporal dependence required to make use of the recurrence of the network. Intuitively, one expects the system to perform better without memory, so that the classification of the current image would not be influenced by the previous input. This assumption was confirmed in both experiments and numerical simulations. Therefore, the recurrent reservoir computer is reduced to a simple feedforward network in this case.

\begin{figure}[b!]
  \centering
  \includegraphics[width=0.5\textwidth]{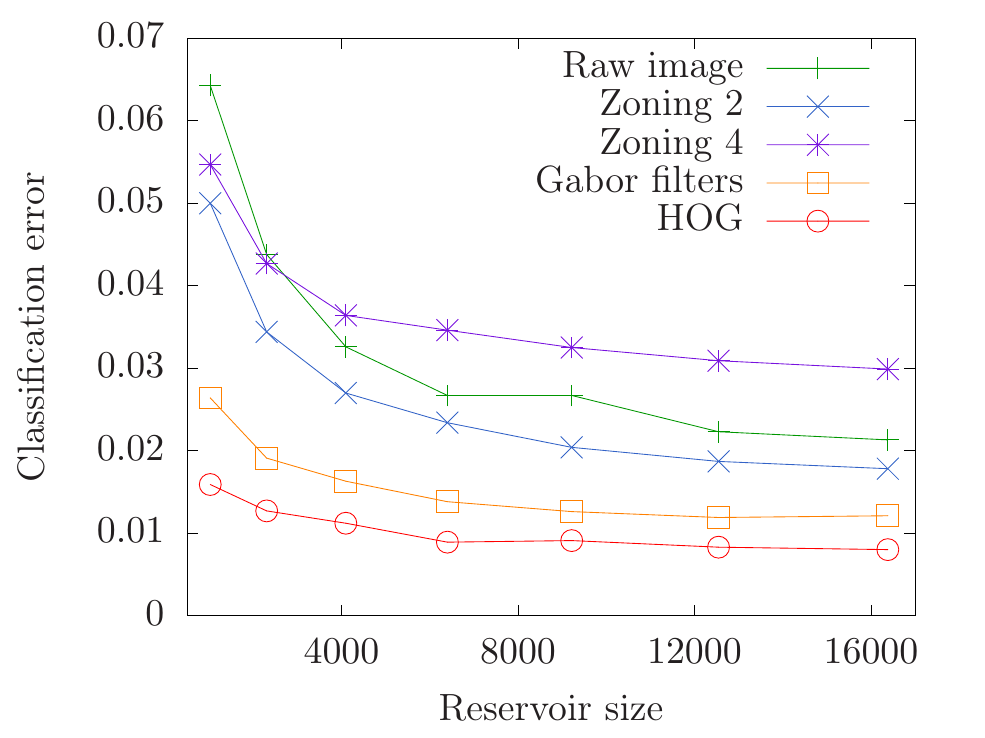}
  \caption{Numerical results obtained with different feature extraction methods and various reservoir sizes. Raw images (unprocessed pixels) are shown for comparison. Zoning 2 technique performs slightly better, while Zoning 4 yields higher classification errors with larger reservoirs. Gabor filters significantly improve performance in comparison to other techniques, but fall short of the HOG algorithm that produced the best results in this study: $0.80\%$ with the largest reservoir of $n=16,384$ nodes.}
  \label{fig:ff}
\end{figure}

Figure \ref{fig:ff} presents a summary of the numerical results obtained with the different feature extraction techniques introduced in Sec. \ref{subsec:feat}. 
We simulated reservoir sizes from 1,024 up to 16,384 nodes. In all cases, the classification error decreases with the number of neurons. 
At first glance, the HOG features produces the lowest classification errors, while raw images and Zoning 4 perform the worst.
We will further discuss each feature extraction methods in the following sections and present the experimental results for comparison.

\subsubsection{Raw images}
\label{subsubsec:res:ff:raw}

\begin{figure}[t]
  \centering
  \includegraphics[width=0.5\textwidth]{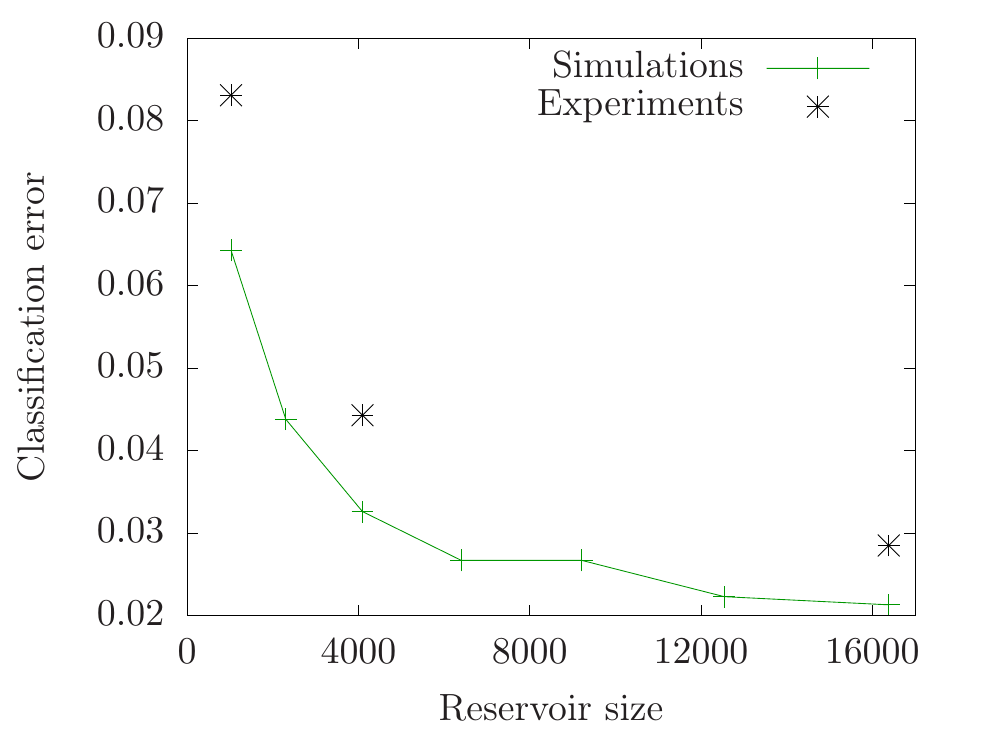}
  \caption{Numerical (green curve) and experimental (black markers) results obtained with the raw images, without any pre-processing or feature extraction. While the experimental results do not match the simulations exactly, they present a similar trend.}
  \label{fig:raw}
\end{figure}

Figure \ref{fig:raw} presents the numerical (green) and experimental (black) results obtained with raw images. 
That is, unprocessed pixels were used as inputs to the reservoir computer. 
We used these results as benchmark for other feature extraction techniques.
In simulations, raw images were classified with a $6.4\%$ error with the smallest reservoir ($n=1,024$ nodes) and $2.1\%$ with the largest reservoir ($n=16,384$ nodes).
In experiments, we obtained a $8.31\%$ and $2.85\%$ error with the smallest and the largest reservoir, respectively.

\subsubsection{Zoning}
\label{subsubsec:res:ff:zon}

Figure \ref{fig:ff} presents the numerical results obtained with both zoning methods -- Zoning 2 with $2\times 2$ windows and Zoning 4 with $4\times 4$ windows.

This technique, as presented in section \ref{subsec:feat}, is the simplest approach considered here, that consists in averaging pixel values over windows of fixed size. The Zoning 2 technique allows to slightly improve the results in comparison with the raw images. Since its effect can be seen as dimensionality reduction, this result indicates that the default $28\times 28$ MNIST images contain some undesirable information that can be removed to improve performance. Zoning 4 technique, on the other hand, performs worse than the raw images, which indicates that too much information is lost when averaging over $4\times 4$ windows. 

Comparing the different curves in Fig. \ref{fig:ff}, one notes that both Zoning techniques provide the lowest classification error improvement in comparison with the other feature extraction techniques. Therefore, we conclude that those two techniques are not suitable for handwritten digit recognition task with reservoir computing. However, they remain of interest as they bring out characteristic spatial scales of the MNIST images.

\subsubsection{Gabor filters}
\label{subsubsec:res:ff:gab}

The case of the Gabor filters perfectly illustrates the main difficulty of the feature extraction approach to classification -- the user has to find the best set of filters for the task at hand, which often requires good understanding of the task under investigation. 

At first, we tried the simpliest approach that consists in computing the energy of the convolution of the filter with the image. In this way, one obtain a single number for each filter. Preliminary simulations with 40 Gabor filters of 8 different orientations and 5 differents lengths have shown a classification error of circa 19\%, which is significantly worse than with raw images (see Fig. \ref{fig:ff}). This result suggests that global features extracted from MNIST images do not contain enough information to distinguish the digits. An intuitive example illustrating the inefficiency of global features is the particular case of separating the digits ``6'' and ``9''. Since the former is very similar to the latter after a rotation by $\pi$, the energy computed from the Gabor filters is approximately the same for both digits, making them indistinguishable from the global features point of view. Others digits, such as ``8'' and ``0'', with comparable curves, also yield energies confusing for the classifier. In order to distinguish these digits, the classifiers needs to know where exactly a particular direction is present in the image, and this is where one needs to use the local features.

At the next stage, we computed local energies by dividing images into smaller windows. This adds another variable parameter to the feature extraction technique : the size of the windows. A sensible choice requires the knowledge of characteristic scales of the MNIST images, and a reasonable first guess can be based on the insights obtained with the zoning techniques, presented in Sec. \ref{subsubsec:res:ff:zon}. Therefore, we computed the energy locally on a $7\times 7$ grid, that is, within $4\times 4$ windows. We fixed 8 directions $\theta$ -- from $0^\circ$ to $157.5^\circ$ by $22.5^\circ$ steps -- and tried different lengths of the filter -- from $2$ to $8$. The lowest classification error of $3.19\%$ was obtained with the filter length $\lambda=7$ pixels, and the smallest reservoir ($n=1024$ nodes). Local energy computation increases the number of features by a factor of $49$ in this case: that is, with 8 Gabor filters, one obtains $392$ features.

At this point, the performance of the reservoir computer is significantly better than with raw images, but the HOG approach remains unmatched (see Fig. \ref{fig:ff}). Therefore, we tried to improve the features obtained with the Gabor filters by taking inspiration from the HOG algorithm. Dalal and Triggs emphasize the importance of block-normalization: without it performance drop by $27\%$ in comparison \cite{dalal2005histograms}. In the next series of numerical simulations, we added $2\times 2$ L2-norm block-normalization. The immediate effect of this operation is the increase of the number of features by a factor of 4 -- from 392 to 1152. Remarkably, we obtain a $26.5\%$-improvement with $8$ Gabor filters of length $\lambda=3$. Simulating the system with filters of different lengths, we obtain the lowest classification error of $2.67\%$ with filters of length $\lambda=5$, and the smallest reservoir ($n=1024$ nodes). This corresponds to a $22.1\%$ improvement gained by adding the block-normalization. 

In summary, Gabor filters could only achieve a circa $19\%$ classification error in the preliminary tests. However,  hand-tailoring the filters to the MNIST database (\textit{i.e.} adding local-energy computation, block-normalization, and setting the filter length to $5$), we managed to reduce the error down to $2.67\%$ with the smallest reservoir ($n=1024$) and $1.21\%$ with the largest reservoir ($n=16384$). We did not manage to outperform the HOG algorithm with Gabor filters in this study, but gained additional insights on the MNIST database and how the components of the HOG algorithm contribute to its performance (in particular, the block-normalisation).

\subsubsection{Histograms of oriented gradients}
\label{subsubsec:res:ff:hog}

The histograms of oriented gradients is a powerful feature descriptor in computer designed for the purpose of object detection \cite{dalal2005histograms}. Similarly to the Gabor filters, the algorithm requires the tuning of a few parameters for better performance. As described in Sec. \ref{subsubsec:feat:hog}, the key parameters are the cell size, the number of bins, and the size of the normalisation blocks. In the case of the MNIST classification task, the generally recommended values for the last two parameters -- 9 bins and $2\times2$ block-normalisation -- produce the best results. As for the cell size, the intuitive guess that smaller cells capture a better representation of the image does not hold here. In fact, the Zoning approach (see Sec. \ref{subsubsec:res:ff:zon}) demonstrated that more information does not always result in better classification.
Preliminary numerical simulations with $4\times4$-pixel cells produced a classification error of $2.1\%$ with the smallest reservoir ($n=1024$). Upon increasing the cell size up to $7\times7$ pixels, we reduced the error down to $1.59\%$ with the same reservoir size. Furthermore, the number of features was cut from $1296$ down to $324$. Further decrease of the number of features by limiting the number of bins down to $4$ (instead of $9$) did not improve performance.

\begin{figure}[t]
  \centering
  \includegraphics[width=0.5\textwidth]{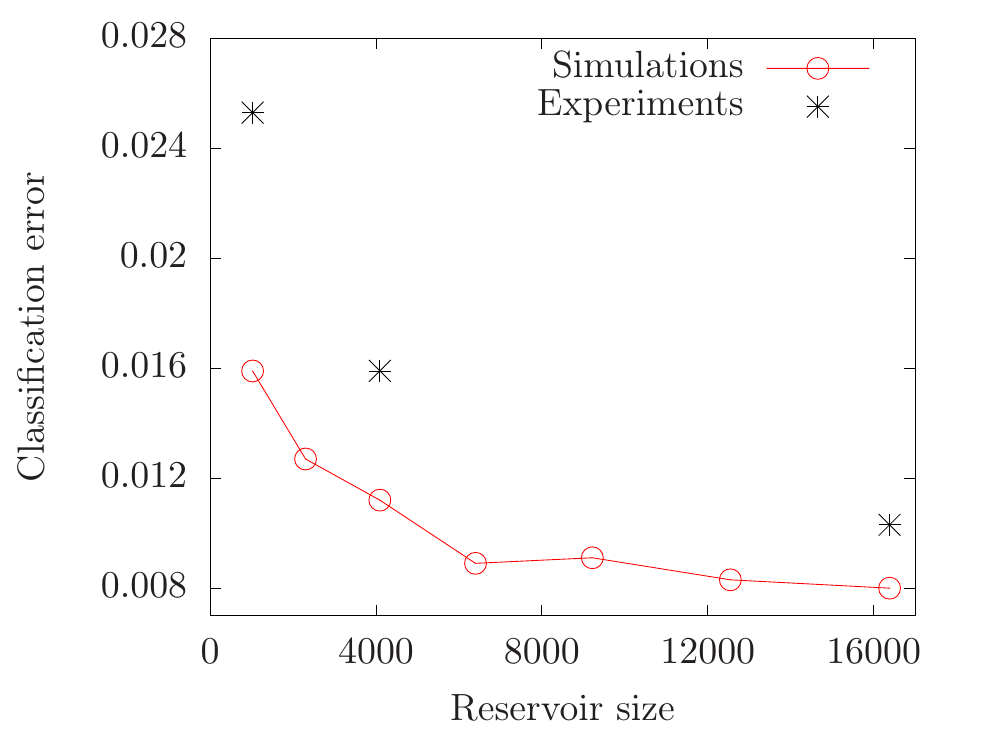}
  \caption{Numerical (red curve) and experimental (black markers) results obtained with the histograms of oriented gradients. 
  Similarly to the raw images, one observes a certain offset with the small and the midsize reservoir. The numerical and experimental performances obtained with the largest reservoir are fairly close, on the other hand.}
  \label{fig:hog}
\end{figure}

Figure \ref{fig:hog} presents the numerical (red) and experimental (black) results obtained with the HOG features and different reservoir sizes.
In simulations, the MNIST digits were classified with a $1.59\%$ error with the smallest reservoir ($n=1,024$ nodes) and $0.8\%$ with the largest reservoir ($n=16,384$ nodes).
In experiments, we obtained a $2.53\%$ and $1.03\%$ error with the smallest and the largest reservoir, respectively.

These results are of the same order of magnitude as the best performance found in the literature, summarized in Tab. \ref{tab:sota}. At the moment of writing this article, the best performance on the MNIST task was obtained with CNNs \cite{wan2013regularization} with numerous approaches reporting  classification errors below $1\%$ (they will not be exhaustively presented here). We selected and listed several alternative methods to CNN, not relying on neural networks, such as SVM, k-means, k-nearest neighbours (KNN), and boosted stumps. Table \ref{tab:sota} shows that our photonic RC outperforms, in numerical simulation, several deep approaches, such as deep Boltzman machines (DBM). Furthermore, the experiment produces lower error than the earlier versions of CNNs, such as LeNet-4\cite{lecun1998gradient}.
One should keep in mind the difference between a shallow neural network, implemented in photonic hardware, and a complex convolutional neural network, implemented on ideal noiseless digital processor. Finally, the reservoir computer significantly improves the performance compared to non-linear and linear classifiers.

\begin{table}
  \centering
  \begin{tabular}{lll}
    \hline
    Authors & Method & Performance \Tstrut\Bstrut\\
    \hline
    Wan et al. \cite{wan2013regularization} & CNN & $0.21\%$ \TstrutL\\
    Wang et al. \cite{wang2016unsupervised} & K-Means & $0.35\%$ \\
    Keysers et al. \cite{keysers2007deformation} & K-NN & $0.52\%$ \\
    Decoste et al. \cite{decoste2002training} & SVM & $0.56\%$ \\
    \textbf{This work (simulations)} & \textbf{Photonic RC} & \textbf{0.8\%} \\
    K\'egl et al. \cite{kegl2009boosting} & Boosted Stumps & $0.87\%$ \\
    Schaetti et al. \cite{schaetti2017echo} & RC & $0.93\%$ \\
    Salakhutdinov et al. \cite{salakhutdinov2009deep} & DBM & $0.95\%$ \\
    \textbf{This work (experiment)} & \textbf{Photonic RC} & \textbf{1.03\%} \\
    LeCun et al. \cite{lecun1998gradient} & LeNet-4 & $1.1\%$ \\
    LeCun et al. \cite{lecun1998gradient} & Non-linear classifier & $3.3\%$ \\
    LeCun et al. \cite{lecun1998gradient} & Linear classifier & $7.6\%$ \BstrutL\\
    \hline
  \end{tabular}
  \vspace{0.25cm}
  \caption{Performance of various state-of-the-art digital approaches compared to our best results on the MNIST database (with $n=16,384$).}
  \label{tab:sota}
\end{table}

\subsection{Full-image recurrent network}
\label{subsec:res:rc}

This second approach is similar to the first one (see Sec. \ref{subsec:res:ff}) in the way that the reservoir computer is presented with the full MNIST image and the features extracted from it. The difference lies in the fact that each subsequent image is input with a fixed delay, so that the temporal dynamics of the reservoir can be exploited to process the information. The transients, i.e. the states of the nodes induced by the input, are recorded and concatenated together into a single reservoir state, used for training. This approach effectively increases the number of trainable parameters, since the size of $W_\text{out}$ is now proportional to the duration of the transients. Consequently, it also lengthens the overall processing and makes the training process more computationally expensive, especially the matrix inversion operation.

The main downside of this technique is the increased number of hyper-parameters of the reservoir. That is, on top of the input gain, the properties of the $W_\text{res}$ matrix (such as the spectral radius, the amplitude of the diagonal elements, and others) have to be optimized to improve the results. As experimental optimization of the hyper-parameters requires long runs in practice, we only investigated this method in numerical simulations.

\begin{figure}[t]
  \centering
  \includegraphics[width=0.5\textwidth]{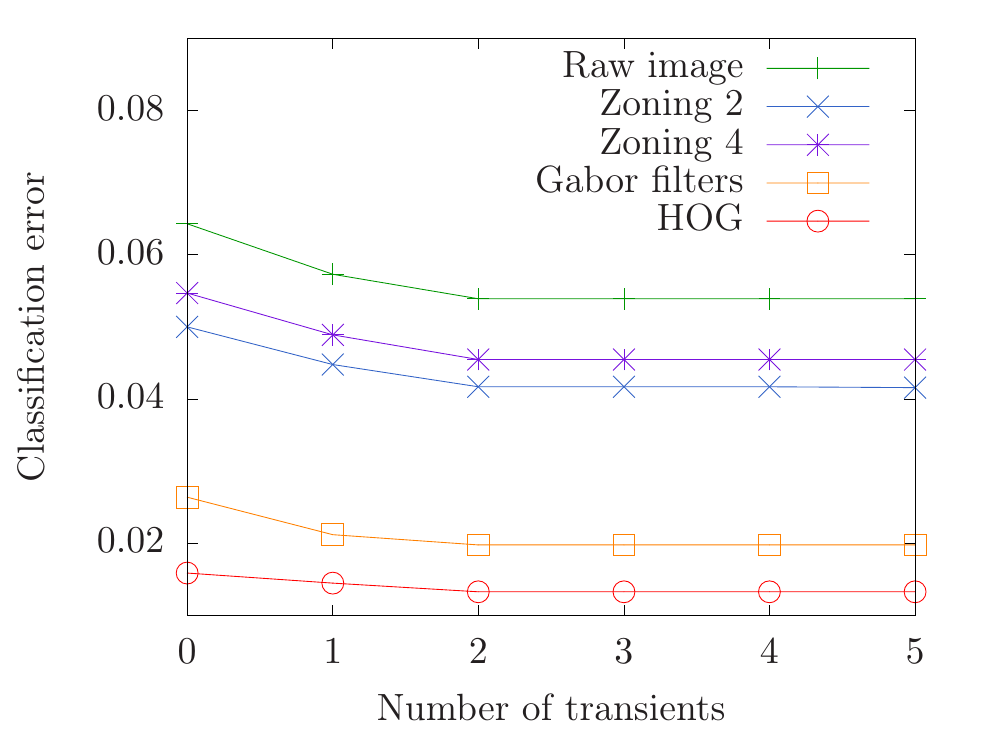}
  \caption{Numerical results obtained with the recurrent network ($N=1024$) approach and various feature extraction methods. Different lengths of transients have been investigated: zero transients corresponds to the feedforward network case, discussed in Sec. \ref{subsec:res:ff}. All curves decrease until two transients and then remain constant.}
  \label{fig:rc_all}
\end{figure}

Figure \ref{fig:rc_all} presents the numerical results obtained with this approach. The classification error is plotted against the number of transients recorded for training using the same feature extraction techniques as in Sec. \ref{subsec:res:ff}. All results were obtained with the smallest reservoir ($n=1024$) and optimized hyper-parameters. The left-most point of each curve corresponds to the feedforward network approach, that is, no transient is recorded and the images are input at each time step $k$. The graph reflects the two major observations produced by this study: (1) the additional transients improve the classification performance for all feature extraction techniques (including raw images), and (2) two transients is the optimal setting for all cases. Table \ref{tab:rc_gains} contains the performance improvements obtained with 2 transients as compared to the feedforward network.

\begin{table}
  \centering
  \begin{tabular}{lccc}
    \hline
    Feature extraction method & Feedforward & Recurrent & Gain \Tstrut\Bstrut\\
    \hline
    Raw images & $0.0643$ & $0.0539$ & $16.2\%$ \TstrutL\\
    Zoning 2 & $0.0500$ & $0.0417$ & $16.6\%$ \\
    Zoning 4 & $0.0547$ & $0.0455$ & $16.8\%$ \\
    Gabor filters & $0.0264$ & $0.0198$ & $25.0\%$ \\
    HOG & $0.0159$ & $0.0133$ & $16.4\%$ \BstrutL\\
    \hline
  \end{tabular}
  \vspace{0.25cm}
  \caption{Performance improvement offered by the recurrent approach with the smallest reservoir ($N=1024$).}
  \label{tab:rc_gains}
\end{table}

\begin{figure}[t]
  \centering
  \includegraphics[width=0.5\textwidth]{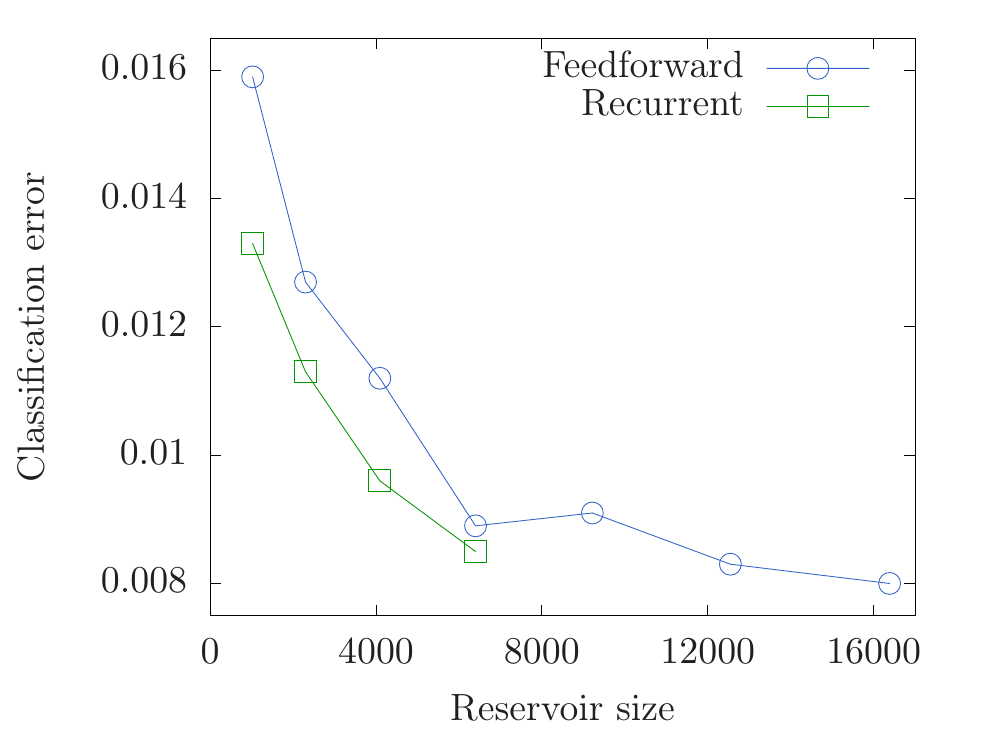}
  \caption{Numerical results obtained with the histograms of oriented gradients, processed by a feedforward network (blue) and a recurrent network (green) with 2 transients. Due to computational complexity, we could only train reservoirs up to $N=6400$ nodes.}
  \label{fig:rc_hog}
\end{figure}

Since the additional transients improve the classification error for the smallest reservoir ($N=1024$), we investigated its effects on larger reservoirs. For simplicity, we only considered the HOG features, since they yield the lowest error rate with the feedforward network (see Sec. \ref{subsubsec:res:ff:hog}). At this stage, we could perceive the increased complexity of training a reservoir with transients, and could only perform numerical simulations up to $N=6400$ nodes on our desktop computers with 32 Gb of RAM. Figure \ref{fig:rc_hog} compares the classification errors obtained with feedforward and recurrent networks. The optimal number of transients was found to be 2, similarly to the previous observations. Table \ref{tab:rc_hog_gains} lists the performance gains for different reservoir sizes.

\begin{table}
  \centering
  \begin{tabular}{lccc}
  \hline
    Reservoir size ($N$) & Feedforward & Recurrent & Gain \Tstrut\Bstrut\\
    \hline
    1024 & $0.0159$ & $0.0133$ & $16.4\%$ \TstrutL\\
    2304 & $0.0127$ & $0.0113$ & $11.0\%$ \\
    4096 & $0.0112$ & $0.0096$ & $14.3\%$ \\
    6400 & $0.0089$ & $0.0085$ & \hfill $4.5\%$ \BstrutL\\
    \hline
  \end{tabular}
  \vspace{0.25cm}
  \caption{Performance improvement offered by the recurrent approach with the histograms of oriented gradients and different reservoir sizes.}
  \label{tab:rc_hog_gains}
\end{table}


In summary, although the recurrent approach increases the number of trainable parameters, lengthens the training process and makes it more computationally expensive, it offers interesting results that have to be taken into account for improving the performance of our reservoir. 

\subsection{Column-wise recurrent mode}
\label{subsec:res:col}

The third approach, inspired by \cite{schaetti2017echo}, uses an explicit temporal encoding of the spatial visual information from the MNIST images in order to activate the recurrent dynamics of the reservoir computer. The idea here is to split the full image into smaller portions and feed them sequentially into the classifier. The separation can be done in various ways: columns or rows (overlapping or adjacent), sliding windows, sub-images, among others. Similarly to \cite{schaetti2017echo}, we consider non-overlapping columns, which transforms each image into 28 inputs of 28 dimensions. Therefore, such temporal encoding reduces the input dimensionality, but increases the processing time proportionally. 

After defining the encoding procedure, the specifics of the reservoir training have to be considered. That is, as each image corresponds to 28 inputs, which in turn give birth to 28 reservoir states, one needs to define how to use those states to train the output weights. We chose to investigate two approaches here:
\begin{itemize}
    \item The first training procedure consists in considering each column as an individual input and train the reservoir computer to output the target digit on every input column forming the corresponding image. Intuitively, one does not expect the system to produce the correct class at the first input column, but assumes the output to converge to the target digit with the last columns of the image. Therefore, while all 28 reservoir states are used for training for each image, the evaluation should only be performed on the basis of the last columns. In this work, we chose to select the last 7 columns and assign the final decision to the most frequent class among the 7 outputs.
    \item The second training method is inspired by \cite{schaetti2017echo} and consists in combining a selection of reservoir states at specific time steps into one, and computing the output weights based on the aggregate reservoir state. This idea is fairly similar to the recurrent network approach (see Sec. \ref{subsec:res:rc}), as it effectively increases the network size by combining several reservoir states into one, and thus expanding the number of trainable parameters.
\end{itemize}

The results of these two approaches are presented in Secs. \ref{subsubsec:res:col:each} and \ref{subsubsec:res:col:comb}, respectively. We used a small reservoir computer ($n=1024$) in all simulation for practical reasons.

\subsubsection{Training with individual reservoir states}
\label{subsubsec:res:col:each}

The first training option appears more challenging for the reservoir computer, since it requires a system with memory long enough to keep all relevant columns of the image. 
In fact, we obtained a relatively high classification error of $21.2\%$. Upon investigation of the reservoir signals we noticed that while the outputs converge, as expected, they often converge to a wrong class. This observation suggests that individual columns do not contain enough relevant information to allow the classifier to make the correct decision.

\subsubsection{Training with aggregation of reservoir states}
\label{subsubsec:res:col:comb}

The second training option alleviates the issue encountered with single columns, as the reservoir computer is trained on an aggregation of several reservoir states activated by their corresponding input columns. This methods also relaxes the memory requirements, as the explicit aggregation of reservoir states no longer forces the network to keep the input information in memory.

The choice of the reservoir states to aggregate for the training is crucial in this approach. That is, training the system on the combination of the first three columns, for instance, is unlikely to produce low classification errors. For this reasons, we started by fixing the hyperparameters of the reservoir computer to reasonably good values (yet not the most optimal), and tested, in numerical simulations, all possible combinations of training with two reservoir states (that is, 378 trials). We found that the combination of the 17-th and 24-th reservoir states yielded the lowest error. Furthermore, we observed that no states below 10-th were chosen in the top ten results. We thus conclude that the earlier reservoir states (from the first to circa the 10-th) do not contain enough information on the digit to allow precise classification, and the later states (from 11-th to the last) should be privileged.
Then, we performed the same series of trials with 3 reservoir states. 
In order to reduce the number of trials from 3276 down to 969, we ignored the reservoir states from 1 to 10, based on the observation above. 
We found the best combination of states to be the 14-th, the 18-th, and the 26-th.
Finally, we tried the combinations of 4 states, and found the optimal one to be the 14-th, the 16-th, the 20-st, and the 24-th.

\begin{figure}[t]
  \centering
  \includegraphics[width=0.5\textwidth]{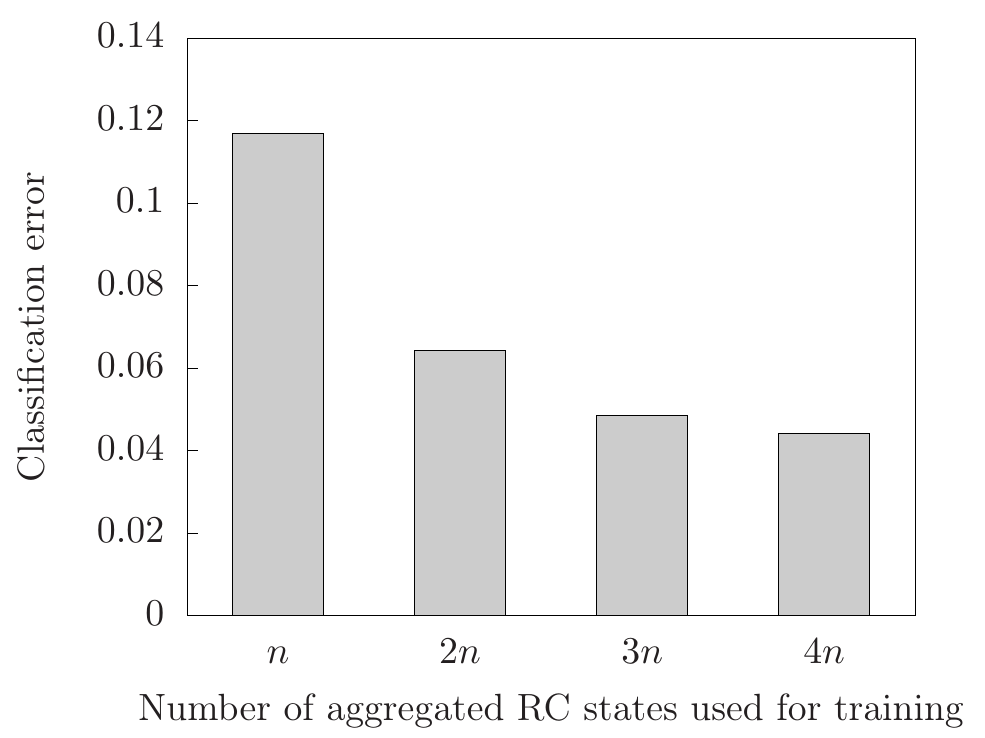}
  \caption{Numerical results obtained with column-wise input data encoding and aggregation of several reservoir states for the training process, with $n=1024$. In particular, training with $n$ corresponds to optimising the readouts weights on the basis of the sole reservoir state, produced after the processing of the 17th column of the image. Conversely, training with $4n$ consists of aggregating four reservoir states, obtained by processing the columns 14, 16, 20, and 24 of the image, for the training process.}
  \label{fig:cols_comb}
\end{figure}

After defining the optimal combinations of reservoir states to aggregate for the training, we kept them fixed, and performed the optimisation of the hyperparameters. Figure \ref{fig:cols_comb} presents the classification errors with optimal training and reservoir settings. For completeness, we also evaluated the scenario with one column, and obtained a $11.7\%$ error with the 17-th column. The best performance of $4.42\%$ was produced with the combination of 4 reservoir states. Compared to the results obtained with raw images, discussed in Secs. \ref{subsubsec:res:ff:raw} and \ref{subsec:res:rc}, this method yields lower errors and thus presents the best approach on raw images among those considered in this study. Nevertheless, the $4.42\%$ classification error is no match for the HOG technique (see Sec. \ref{subsubsec:res:ff:hog}) or the state-of-the-art.

\section{Conclusion}
\label{sec:ccl}

In summary, we report two major results in this paper. First, we present a large-scale experimental photonic reservoir computer, based on off-the-shelf components, capable of implementing neural networks up to $16,384$ nodes. Second, the computational power offered by our large dynamical system allows us to tackle an important task in computer vision: the recognition of handwritten digits. To this end, we study three conceptually different approaches of applying our reservoir computer to this task. First, we exploit the reservoir in a memoryless, feedforward mode of operation paired with popular handcrafted feature-extraction techniques (Zoning, Gabor filters, and HOG) prior to feeding the input layer of the photonic reservoir. The second mode of operation consisted in using the temporal dynamics of the network and hence benefit from its internal memory enabled by recurrent connections, still paired to feature extraction. Third, we used the temporal dynamics of our reservoir computer with a time-dependent signal, created by splitting the raw MNIST images into individual columns, thus creating temporal dependence and stimulating the recurrence of the system. We report a $0.8\%$ classification error in numerical simulations and $1.03\%$ error in experiments, obtained with the feedforward mode of operation using the HOG feature extraction technique. We also demonstrated in numerical simulations the benefit of the recurrence mode of operation allowing for relative gain improvement ranging from $4.5\%$ to $16.4\%$ for the smallest reservoir in our study and the superiority of preserving spatial correlation and feature extraction compared to transforming the 2D MNIST images into multiple 1D signals. In this paper, we have shown that large-scale photonic reservoir computer can perform with level of performance comparable to the best digitally-implemented neuro-inspired currently available and hence represent a meaningful  alternative to these methods, hence opening more in-depth research endeavours on exploiting photonic reservoir computer for advanced image processing.

In perspective, our work opens several directions for future research. The handcrafted features could be replaced by transfer learning, i.e. by automatically synthesised features from the first layers of a convolutional neural network trained on the MNIST dataset \cite{olivas2009handbook}. Furthermore, since the reservoir computer yields competitive results on handwritten digit recognition, one could increase the complexity of the task and apply the system to visual object recognition using e.g. the CIFAR-10 dataset \cite{krizhevsky2009learning}.
The present work is thus the first step in what we hope should be a very fruitful line of investigations.



%

%

\section*{Acknowledgment}

This work was supported by the AFOSR (grants No. FA-9550-15-1-0279 and FA-9550-17-1-0072), the R\'egion Grand-Est. The authors gratefully acknowledge Dr. Daniel Brunner for the insightful scientific exchanges and discussions related to spatiotemporal photonic reservoir computers.

\ifCLASSOPTIONcaptionsoff
  \newpage
\fi



\bibliographystyle{IEEEtran}
\bibliography{refs.bib}
%

%

\begin{IEEEbiography}[{\includegraphics[width=1in,height=1.25in,clip,keepaspectratio]{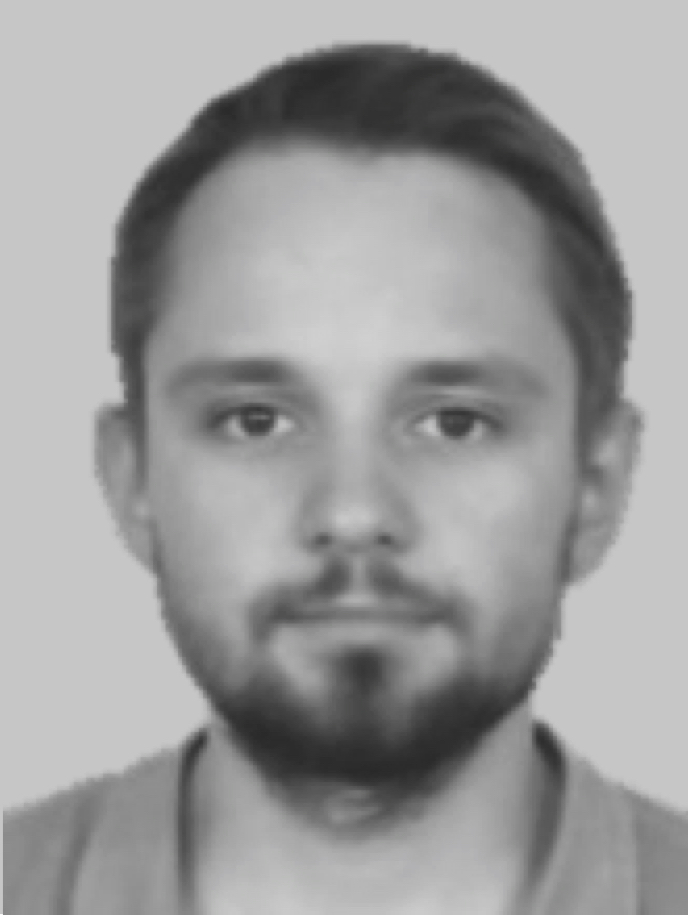}}]{Piotr Antonik} was born in Minsk, Belarus, in 1989. He graduated in physics from the Universit\'e libre de Bruxelles (Brussels, Belgium) in 2013 and defended his PhD in 2017, under the direction of Prof. S. Massar. During his PhD, he studied implementations of machine learning methods in photonic systems. From 2017 to 2018 he was a Post-Doctoral researcher with the LMOPS EA-4423 and the Chair in Photonics at CentraleSupélec in Metz, France. In October 2018, he obtained a permanent position of Associate Professor at CentraleSupélec, Metz Campus, with the LMOPS EA-4423 and the Chair in Photonics. He continues his research activities in machine learning and photonics, with applications in e.g. biomedical imaging and telecommunications. To this day, Dr. Antonik published 6 articles in peer-reviewed journals and presented 16 contributions (talks and posters) to international conferences. His PhD thesis won the Springer Theses Award and was published in the Springer Theses collection.
\end{IEEEbiography}

\begin{IEEEbiography}[{\includegraphics[width=1in,height=1.25in,clip,keepaspectratio]{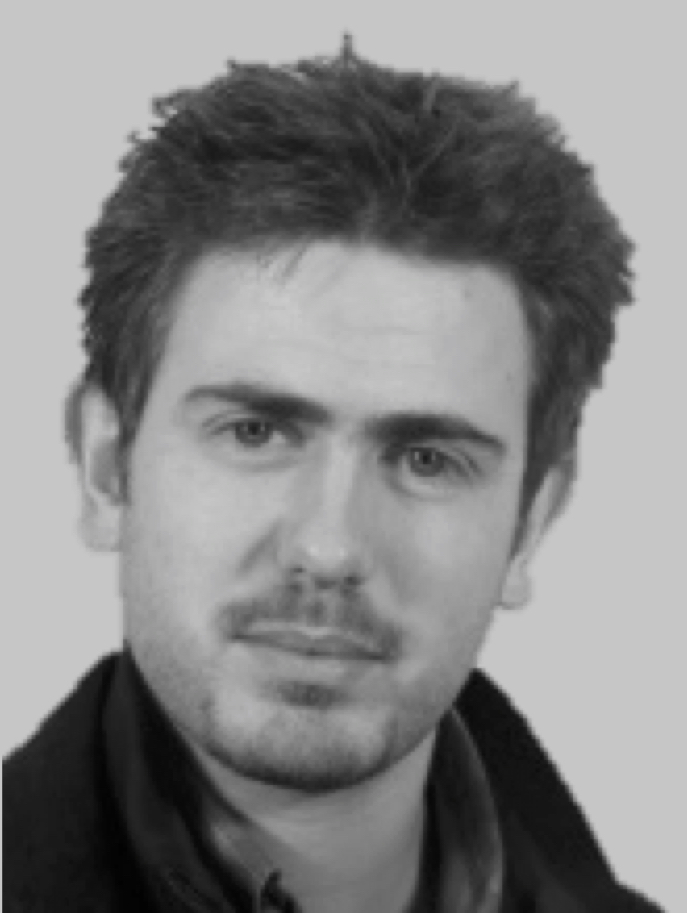}}]{Nicolas Marsal}
was born in Woippy, France. He received his Master of Physics from Universit\'{e} de Lorraine (France) in 2007. In 2010, he defended his PhD in Physics at CentraleSup\'{e}lec. From 2011 to 2012, he was a Postdoctoral Fellow in the Nonlinear Optics department at Georgia Tech Lorraine (France). Since December 2012, he got a permanent position of assistant professor at CentraleSup\'{e}lec. His research activities are mainly focused on spatio-temporal dynamics of nonlinear photonic systems, propagation and interaction of non-conventional optical beams and machine learning at the physical layer for innovative applications in neuro-inspired information processing and cognitive computing. His research is conducted within the framework of the LMOPS EA-4423 Laboratory, a joint unit between CentraleSup\'{e}lec and the Universit\'{e} de Lorraine (France). Dr Marsal has authored and co-authored 23 publications in international peer-reviewed journals and presented more than 50 contributions in international conferences.
\end{IEEEbiography}


\begin{IEEEbiography}[{\includegraphics[width=1in,height=1.25in,clip,keepaspectratio]{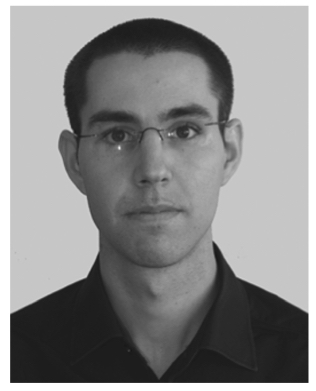}}]{Damien Rontani}
was born in Clamart, France. He received his Master of Science in Electrical Engineering (Sup\'{e}lec Curriculum) and a Master in Information Science, Energy and Systems from CentraleSup\'{e}lec in Universit\'{e} Paris-Saclay (France) in 2004 and 2006, respectively. In parallel, he received his Master of Science in Electrical and Computer Engineering from the Georgia Institute of Technology (Georgia Tech, USA) in 2005. In 2011, he defended his PhD in Photonics at CentraleSup\'{e}lec and his PhD in Electrical and Computer Engineering at Georgia Tech. From 2011 to 2013, he was a Postdoctoral Fellow in the Physics Department at Duke University (USA). Since December 2013, he has been an assistant professor at CentraleSup\'{e}lec, where he conducts leading-edge research in nonlinear dynamics of photonics systems, nonlinear optics, and machine learning at the physical layer for innovative applications in neuro-inspired information processing and cognitive computing. His research is conducted within the framework of the LMOPS EA-4423 Laboratory, a joint unit between CentraleSup\'{e}lec and the Universit\'{e} de Lorraine (France). In his is young academic career, he has been the recipient of an IBM Faculty Award in Cognitive Computing in 2015 and a JSPS Fellowship for Overseas Researchers from the Japanese Society for the Promotion of Science in 2016. He has authored and co-authored 24 publications in international peer-reviewed journals and more than 50 contributions in international conferences.
\end{IEEEbiography}




\end{document}